\documentclass{report}

\usepackage{arxiv}

\usepackage[utf8]{inputenc} 
\usepackage[T1]{fontenc}    
\usepackage{hyperref}       
\usepackage{url}            
\usepackage{booktabs}       
\usepackage{amsfonts}       
\usepackage{nicefrac}       
\usepackage{microtype}      
\usepackage{lipsum}		
\usepackage{graphicx}
\usepackage{doi}
\usepackage{placeins}
\usepackage[backend=biber, style=authoryear]{biblatex}
\title{A review on different techniques used to combat the non-iid and heterogeneous nature of data in FL}

\author{ \href{https://orcid.org/0000-0000-0000-0000}{\includegraphics[scale=0.06]{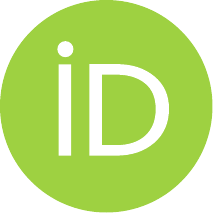}\hspace{1mm}Venkataraman Natarajan Iyer} \\
	Mechanical and Aerospace Engineering\\
	Nanyang Technological University
\\
	Singapore \\
	\texttt{venkataramannatarajan2001@gmail.com
 } \\
}
\date{}

\hypersetup{
pdftitle={A template for the arxiv style},
pdfsubject={q-bio.NC, q-bio.QM},
pdfauthor={David S.~Hippocampus, Elias D.~Striatum},
pdfkeywords={ },
}

\begin{document}
\maketitle
\begin{abstract}
Federated Learning (FL) is a machine-learning approach enabling collaborative model training across multiple decentralized edge devices that hold local data samples, all without exchanging these samples. This collaborative process occurs under the supervision of a central server orchestrating the training or via a peer-to-peer network. The significance of FL is particularly pronounced in industries such as healthcare and finance, where data privacy holds paramount importance.

However, training a model under the Federated learning setting brings forth several challenges, with one of the most prominent being the heterogeneity of data distribution among the edge devices. The data is typically non-independently and non-identically distributed (non-IID), thereby presenting challenges to model convergence. This report delves into the issues arising from non-IID and heterogeneous data and explores current algorithms designed to address these challenges.  
\end{abstract}


\section{Introduction}
Traditional machine learning involves the collection of extensive data, followed by pre-processing, enabling the training of large-scale models in a centralized manner. The pre-processing step is crucial, enhancing data quality and facilitating the model's ability to learn patterns for accurate predictions. In centralized machine learning scenarios, where data is sourced from a single, well-controlled environment, the assumption of Independent and Identically Distributed (IID) data is often reasonable.

The term Federated learning was introduced by Google in 2016 as part of a paper that aimed to solve the problem of training a centralized machine learning model, specifically Google Keyboard, from data distributed among millions of clients, specifically mobile phones. Training in Federated learning can be done under two different settings, namely, Centralized Federated learning and Decentralized Federated learning. 

\begin{figure}[!htbp]
    \centering
    \includegraphics[width=60mm]{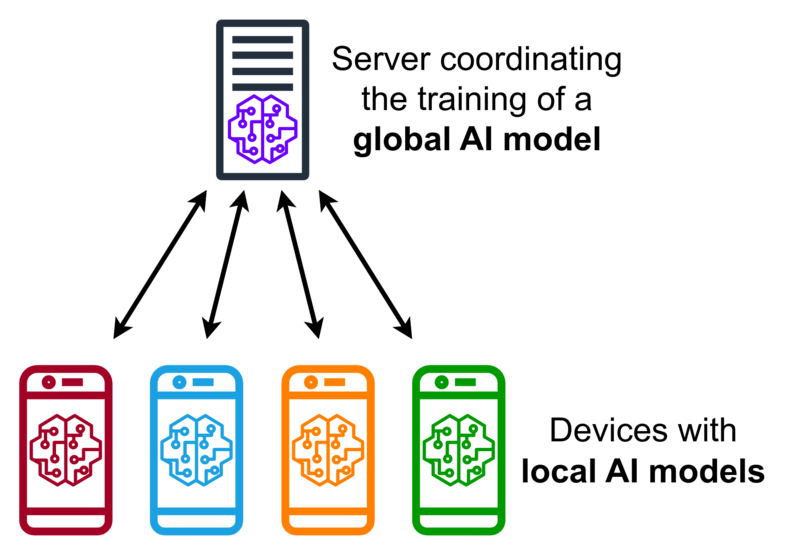}
    \caption{Federated learning}
    \label{fig:Federated learning}
\end{figure}
\FloatBarrier

\textbf{Centralized Federated learning:} In the Centralized Federated learning setting, a central server is used to orchestrate and coordinate the training process amongst all the clients or edge devices. This involves selecting the participating clients for the current round of training, aggregating the global model based on the local model updates from each participating client, and broadcasting the updated model to the edge devices.

\textbf{Decentralized Federated learning:} In the Decentralized Federated learning setting, the edge devices communicate and share the global model. This setting prevents a single point of failure by alleviating the need to have a central server that orchestrates the training. However, this kind of network topology may have an impact on the accuracy of the trained model. 

In Federated Learning, the pre-processing of all data is unfeasible since data samples reside locally on edge devices. The data from these edge devices may be sampled from varying sources and hence have different distributions. Consequently, the assumption of IID data is no longer tenable. The training data in Federated Learning becomes heterogeneous and non-IID.

Heterogeneous data encompasses datasets where samples or data points originate from different distributions or possess diverse characteristics. For instance, in medical data collected from various hospitals, each facility may adopt distinct recording methods, introducing differences in data distribution. Heterogeneous data exhibits variability in features, formats, or sources, presenting a more challenging modeling task.

Non-IID data implies that samples or data points in a dataset are not independent and identically distributed. In other words, the assumption that each data point is drawn from the same underlying distribution breaks down. For example, in a recommendation system, changing user preferences over time can render the data non-IID. Non-IID data often arises due to either temporal dependencies or spatial correlations between data points.

Heterogeneous and non-IID data bring forth challenges in training a model under a Federated learning setting. Some of the challenges involved are listed below:

\textbf{Model Heterogeneity:} Heterogeneous data involves data sampled from clients having varying distributions. This makes it challenging to train a global model that performs well across all edge devices and clients.

\textbf{Convergence challenges:} Heterogeneous and non-IID data can lead to a slower convergence of the model and in some cases may also result in model divergence. Some of the existing Federated learning algorithms such as FedAvg which involves the averaging of the individual client model updates may result in significant model divergence from the optimum, especially if the training data of one of the clients has a distribution vastly different from the other clients.

\textbf{Sampling Bias:} Non-IID data may result in models that are biased toward specific subpopulations. It's crucial to address sampling bias to ensure fairness and generalization across diverse user groups.

\textbf{Adaptability Issues:} The data from various clients may change over time. Ensuring that the global model can quickly adjust to local changes without comprising the overall performance of the model may be a challenge. 

\textbf{Robustness:} Building models that are robust to variations in data distribution and capable of generalizing well across different data sources is a key challenge in federated learning with heterogeneous and non-IID data.

Some of the other challenges include \textbf{communication overhead}, \textbf{privacy concerns}, \textbf{security risks} and \textbf{resource constraints}.

Some notable Federated learning algorithms considered as state-of-the-art are \textbf{FedAvg}, \textbf{FedProx}, \textbf{FedAdam}, \textbf{FedMA}, \textbf{FedMeta}, \textbf{FedDP}, \textbf{FedNova} and \textbf{SCAFFOLD}. The next section provides a brief introduction to some of the above algorithms before delving into the mathematical definitions of heterogenous and non-IID data and their impact on Federated learning algorithms. 

\section{Background Review}
\textbf{FedAvg (Federated Averaging):} \textbf{FedAvg} is a foundational Federated learning algorithm where model updates from different devices are averaged to update a global model.
\begin{center}
   ${\theta_{t + 1}}$ = ${\frac{1}{K}\sum_{k=1}^{K}\theta_{t}^{(k)}}$ 
\end{center}
where ${\theta_{t + 1}}$ is the updated global model, ${K}$ is the total number of clients and ${\theta_{t}^{(k)}}$ is the model update from device ${k}$ at iteration ${t}$.

\textbf{FedProx (Federated Learning with Proximal Term):} \textbf{FedProx} extends \textbf{FedAvg} by introducing a proximal term in the optimization objective to mitigate the impact of non-IID data distribution. This modified objective function is used by the clients to train on their local datasets. The local model updates are aggregated using the \textbf{FedAvg} by the central server. The modified objective function is given below:
\begin{center}
   ${min_{\theta}(f_{k}(\theta) + \frac{\lambda}{2}||\theta - \theta_{old}||^{2}})$ 
\end{center}
where ${f_{k}(\theta)}$ is the local loss function of device ${k}$ with model parameters ${\theta}$, ${\lambda}$ is a regularization parameter and ${\theta_{old}}$ is the previous global model.

\textbf{FedNova (Federated Learning with Novelty and Variance Adaptation):} FedNova is designed to improve convergence in federated learning, especially in the presence of non-IID data. It adapts the learning rate based on the novelty and variance of local model updates. The local model update formula is given below:
\begin{center}
   ${\theta_{t + 1} = \theta_{t} - \eta_{t + 1} . \nabla F(\theta_{t})}$ 
\end{center}

\begin{center}
   ${\eta_{t + 1} = \alpha . max(\sqrt{\frac{D_{t}}{D_{ref}}}, \beta)}$ 
\end{center}
where ${D_{t}}$ is the variance of the local model updates, ${D_{ref}}$ is a reference variance, ${\alpha}$ is a scaling factor, ${\beta}$ is a threshold, ${\theta_{t}}$ is the current global model and ${\nabla F(\theta_{t})}$ is the gradient of the objective function concerning the model parameters.

The mathematical definitions of heterogeneous and non-IID data are described below:

\textbf{Heterogeneous Data:} Given a dataset ${D}$ with samples ${{x_{1}, x_{2}, ...., x_{n}}}$, and each sample having ${m}$ features, the dataset is heterogeneous if the samples are drawn from different distributions: ${P(x_{i}) \neq P(x_{j})}$ for some ${i \neq j}$. This inequality indicates that the probability distributions of different samples ${x_i}$ and ${x_j}$ are not equal.

\begin{figure}[!htbp]
    \centering
    \includegraphics[width=90mm]{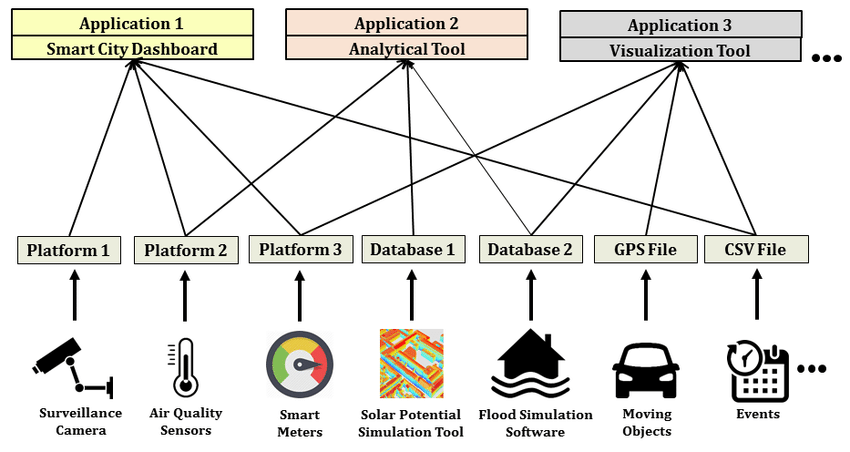}
    \caption{Heterogeneous data}
    \label{fig:Heterogeneous data}
\end{figure}
\FloatBarrier

\textbf{Non-IID Data:} Given a dataset ${D}$ with samples ${{x_{1}, x_{2}, ...., x_{n}}}$, and each sample having ${m}$ features, the IID assumption is that the samples are independent and identically distributed: ${P(x_{1}, x_{2}, ...., x_{n}) = P(x_{1}).P(x_{2})....P(x_{n})}$. Non-IID data violates this assumption; there are dependencies between samples, and their joint distribution doesn't factorize as shown above.

\begin{figure}[!htbp]
    \centering
    \includegraphics[width=60mm]{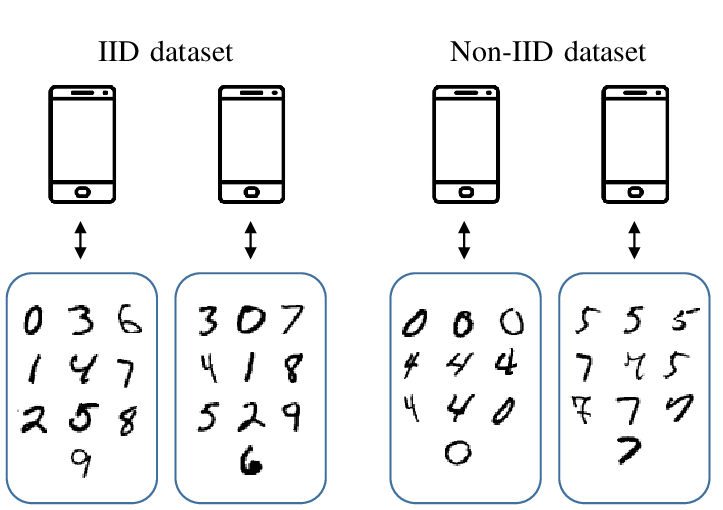}
    \caption{IID vs non-IID data}
    \label{fig:IID vs non-IID data}
\end{figure}
\FloatBarrier

Mathematical techniques can be employed to address heterogeneous and non-IID data. For heterogeneous data, domain adaptation techniques may be employed to align the distributions of different sources. For non-IID data, models need to consider dependencies, potentially using techniques like recurrent neural networks for temporal dependencies or spatial models for spatial correlations.

Few authors have explored the impact of heterogeneous and non-IID data on existing Federated learning algorithms such as \textbf{FedAvg}, \textbf{FedProx}, \textbf{SCAFFOLD}, \textbf{FedNova}, etc. This report examines the work done in \cite{hsu_2019_measuring} and \cite{federated}, summarizing the significant findings by the corresponding authors.

In \cite{hsu_2019_measuring}, the author explored the effects of non-IID data distribution for federated visual classification. The Dirichlet distribution was used to distribute the CIFAR-10 dataset consisting of 10 classes to 100 clients, each having 500 images. The model architecture details are as follows. The same CNN architecture as in \cite{mcmahan_2017_communicationefficient} was used. A weight decay of ${0.004}$ was used and no learning rate decay was applied. A client batch size of ${64}$ was used. The training was carried out with local epochs of ${{1, 5}}$, and with a participating fraction of ${{0.05, 0.1, 0.2, 0.4}}$, where the participating fraction represents the fraction of clients taking part in the federated learning process, involving local training and sharing of local model updates. The training was done for a total of ${10,000}$ communication rounds, where each communication round refers to a cycle in which participating clients collaboratively update the global model.

The observations from the above-described training process are summarised below:
\begin{itemize}
    \item Non-IID data leads to relatively poor classification performance.
    \item Increasing participation fraction has diminishing gains, especially for IID data.
    \item Having fewer epochs does not necessarily increase accuracy for non-IID data.
    \item Training is more volatile with fewer participating clients and also fails to converge after ${10000}$ rounds for non-IID data.
    \item Convergence may not occur with a lower participating fraction for non-IID data.
    \item Convergence may not occur with a higher number of epochs for non-IID data.
\end{itemize}

In \cite{federated}, the author proposed comprehensive data partitioning strategies to cover the typical non-IID data setting in Federated learning. Experiments were conducted to evaluate state-of-the-art Federated learning algorithms under various non-IID data settings.

The comprehensive data partitioning strategy consisted of a total of ${6}$ strategies, which include ${3}$ different types of skews that can be observed in non-IID data. These include \textbf{label skew}, \textbf{feature skew}, and \textbf{sample skew}. The different data partitioning strategies are summarised below:

\textbf{Label distribution skew:} The label refers to the output or the desired prediction for a given input. Label distribution skew refers to an unequal distribution of labels or classes amongst the participating clients. Label distribution skew can be simulated using two different techniques as below.

\textbf{\textit{Quantity-based label imbalance:}} In this technique, different sets of labels were randomly assigned to the clients. The samples of each label were then randomly and equally divided amongst the parties that owned the labels. There may be overlap in the different sets of labels owned by various parties, however, there is no overlap between the samples of different parties.

\textbf{\textit{Distribution-based label imbalance:}} In this technique, each party was allocated a proportion of the samples of each label according to Dirichlet distribution.

\begin{figure}[!htbp]
    \centering
    \includegraphics[width=60mm]{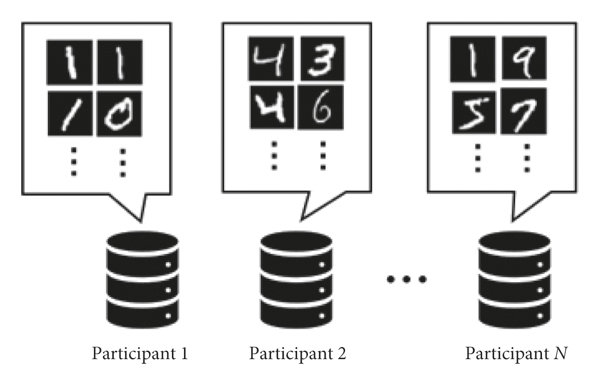}
    \caption{Label distribution skew amongst participants}
    \label{fig:Label distribution skew amongst participants}
\end{figure}
\FloatBarrier

\textbf{Feature distribution skew:} In feature distribution skew, the feature distributions ${P(x_{i})}$ vary across parties. Feature distribution skew can be simulated using three different techniques as below

\textbf{\textit{Noise-based feature imbalance:}} In this technique, the entire dataset was divided into multiple parties randomly and equally. Different levels of Gaussian noise were added to the local dataset of each party to achieve different feature distributions.

\textbf{\textit{Synthetic feature imbalance:}} In this technique, a synthetic feature imbalance federated dataset is created by distributing data points in a cube, which is partitioned into 8 parts by three different planes, wherein each part contains data points of a particular label. A subset for each party is then allocated from two parts which are symmetric of ${(0, 0, 0)}$. This results in varying feature distribution amongst parties.

\textbf{\textit{Real-world feature imbalance:}} This technique is especially usable when the data is collected from different sources. Such data tends to have a natural variance in its feature distribution. Data points can be distributed such that each party has data collected from different sources, thus resulting in a feature imbalance amongst the parties.

\textbf{Quantity skew:} In quantity skew, each party was allocated with a varying size of the local dataset. This size was determined using the Dirichlet distribution. 

\begin{figure}[!htbp]
    \centering
    \includegraphics[width=60mm]{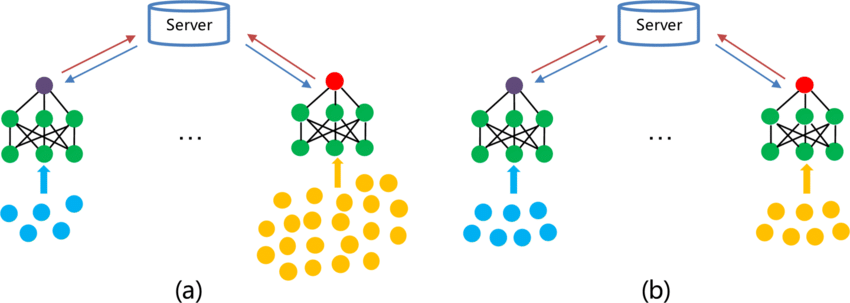}
    \caption{Quantity distribution skew amongst participants}
    \label{fig:Quantity distribution skew amongst participants}
\end{figure}
\FloatBarrier

The observations from the evaluation of state-of-the-art Federated learning algorithms under the above-described data partitioning strategies are summarised below:
\begin{itemize}
    \item Label skew has the most impact on all the existing algorithms.
    \item All algorithms perform relatively worse when each party has only one label.
    \item \textbf{FedProx} performs the best with label and quantity skew.
    \item \textbf{FedProx} has the same convergence speed as \textbf{FedAvg}.
    \item \textbf{FedNova} and \textbf{SCAFFOLD} are relatively unstable, and SCAFFOLD does not work effectively under partial participation.
    \item \textbf{FedProx} has higher computational costs and \textbf{SCAFFOLD} has higher communication costs.
    \item Data distribution that exhibits mixed types of skew impacts the accuracies of all algorithms.
\end{itemize}

\section{Techniques to combat heterogeneous and non-IID data}
This section will examine the work done by authors \cite{tao_2020_ensemble}, \cite{a}, and \cite{MKT}, which explores techniques to combat heterogeneous and non-IID data distribution during Federated learning.

\subsection{Ensemble Distillation for Robust Model Fusion in Federated Learning}
In \cite{tao_2020_ensemble}, the author proposes an ensemble learning approach to achieve knowledge distillation. Ensemble learning methods involve combining the predictions of multiple machine learning models to improve overall performance. Knowledge distillation is a technique in machine learning where a small, compact model (student model) is trained to replicate the behavior of a larger, more complex model (teacher model). The proposed algorithm is called \textbf{Federated Distillation Fusion (FedDF)}. The algorithm involves the following steps. In each communication round, a set of participating clients undergo training to update their local models based on their local data points. After the local training of each participating client is completed, \textbf{FedAvg} is used to initialize the fusion model. A mini-batch of samples ${d}$ is sampled either from an unlabeled dataset or a generated dataset. This mini-batch of data is used as training data to distill the ensemble of teacher models, which consists of the local model of each participating client, to a single server student model. The averaged logits of all the client teachers are used to train the initialized fusion student model on the server. Kullback-Leibler divergence is used as the loss function.  

For two discrete probability distributions ${P}$ and ${Q}$, the ${KL}$ divergence is described below
\begin{center}
   ${D_{KL}(P || Q) = \sum_{i}P(i)log(\frac{P(i)}{Q(i)})}$
\end{center}

The model update equation for the distillation of the ensemble of teacher models is described below:
\begin{center}
   ${x_{t,j} := x_{t,j-1} - \eta\frac{\partial KL(\sigma(\frac{1}{|S_{t}|} \sum_{k \in S_{t}} f(\hat{x}^{k}_{t}, d)), \sigma(f(x_{t,j-1}, d)))}{\partial x_{t,j-1}}}$
\end{center}
where ${\hat{x}^{k}_{t}}$ represents the trained local client model, ${\sigma}$ represents the softmax function, ${d}$ represents the mini-batch of samples, ${S_{t}}$ represents the random subset of clients, and ${x_{t,j}}$ represents iteration ${j}$ in the model fusion update of the ${t^{th}}$ communication round. 

The \textbf{FedDF} algorithm was benchmarked against state-of-the-art Federated learning algorithms such as \textbf{FedAvg}, \textbf{FedProx}, \textbf{FedAvgM}, \textbf{FedMA}. The conditions for the training and evaluation of the performance of the \textbf{FedDF} algorithm are as follows. The algorithm was evaluated using various architectures such as \textbf{ResNet}, \textbf{VGG}, \textbf{ShuffleNETV2}, and \textbf{DistilBERT}. There was no learning rate decay, no momentum acceleration, and no weight decay used. CIFAR-100 was used as the distillation dataset for \textbf{FedDF}.

The observations from the performance of the \textbf{FedDF} algorithm are summarised below
\begin{itemize}
    \item \textbf{FedDF} performs better in terms of accuracy on the test dataset than \textbf{FedAvg}, especially with increased local epochs and a more heterogeneous and non-IID dataset.
    \item Number of communication rounds required to reach the target test accuracy decreases with an increase in the number of local epochs and the fraction of clients being trained. This is the ideal training condition in Federated learning.
    \item \textbf{FedDF} performs better in the absence of normalization techniques like Batch normalization.
    \item \textbf{FedDF} has faster convergence than \textbf{FedAvg} on Natural language processing (NLP) tasks.
\end{itemize}

\subsection{A Robust Aggregation Approach for Heterogeneous Federated Learning}
In \cite{a}, the author proposes a method to aggregate the client model updates by weighing the local client model updates not only on the amount of data but also based on the number of classes or labels each client's local dataset had when trained locally. The proposed method is called \textbf{Federated labels (FedLbl)}. The method divides the local models into two groups based on the heterogeneity level of the data the model has trained on. The heterogeneity level is interpreted based on the number of classes or labels the model is trained on. The heterogeneity level incorporates the volume and variance of the training data of each local client.

If the ${kth}$ client's number of labels is lesser than a predefined threshold, the local model of the client is added to the group ${Z}$, else it is added to the group ${M}$. The global model is optimized based on the following equation:
\begin{center}
    ${w^{t+1} = \sum^{K}_{k=1}(1-\alpha)(\frac{n_{k}}{n}) + \alpha(\frac{\nu}{M})w^{t+1}_{k}}$, if ${c_{k} \in M}$

    ${w^{t+1} = \sum^{K}_{k=1}(1-\alpha)(\frac{n_{k}}{n}) + \alpha(\frac{(1-\nu)}{Z})w^{t+1}_{k}}$, if ${c_{k} \in Z}$
\end{center}

The parameter ${\alpha}$ controls the weight given to the number of labels/classes as opposed to the volume of data that the concerned local client was trained on. The parameter ${\nu}$ controls the preference given to clients whose models were trained with more labels/classes.  

The performance of the algorithm on various datasets such as MNIST, Fashion-MNIST, and CIFAR-10, with various model architectures was observed. The \textbf{FedLbl} algorithm outperforms other algorithms like \textbf{FedAvg}, and \textbf{FedSGD} in metrics such as test dataset accuracy and loss function convergence.

\subsection{Decentralized federated learning via mutual knowledge transfer}
In \cite{MKT}, the author proposes a novel mutual knowledge transfer algorithm called \textbf{Def-KT} in a decentralized federated learning setup. Traditional federated learning involves the participating clients sending their model updates to the central server after each round for updating the global model. The potential problems with a central server include a single point of failure and additional infrastructure requirements for scaling. The proposed peer-to-peer federated learning is a variation of federated learning where the participating devices or nodes in the federated learning system communicate directly with each other in a peer-to-peer fashion, without relying on a central server for coordination. This approach introduces a decentralized and distributed element to the federated learning process.

Mutual knowledge transfer is a process in which two student neural networks learn simultaneously and collaboratively, teaching each other when being trained on the same data. Initializing these neural networks differently can lead to the neural networks learning different knowledge.

The details of the implementation of the \textbf{Def-KT} algorithm are summarised below
\begin{itemize}
    \item In every iteration, ${Q}$ out of ${K}$ clients are selected for training with ${(2Q << K)}$. The ${Q}$ clients train locally on their datasets. After the training, each of the ${Q}$ clients selects a client from another set of ${Q}$ clients to share the model with. The two sets of ${Q}$ clients do not overlap.
    \item The new set of ${Q}$ clients with whom the model is shared, use their local weights and the shared weights to train simultaneously on their local dataset with an interdependent loss function. The shared weights and local weights are updated simultaneously during the training process.
    \item The updated shared weight is set as the new weight.
    \item The loss function used for training is the summation of the Kullback-Leibler divergence function between the predictions of the two models and the Euclidean distance between the prediction of the model being trained and the ground truth.
\end{itemize}

The two weights being trained simultaneously are the shared weights ${w_{k}}$ of the initial set of ${Q}$ clients after their training, and the local weights ${w_{k+Q}}$ of the second set of ${Q}$ clients with whom the weights are shared. The local training data of each of the second set of ${Q}$ clients is split into mini-batches of size ${B}$ for training the weights. The weight update equations are described below:
\begin{center}
    ${P_{1} = model(B, w_{k})}$
    
    ${P_{2} = model(B, w_{k+Q})}$

    ${w_{k} := w_{k} - \eta_{1}\frac{\partial Loss_{1}(w_{k}, B, P_{2})}{\partial w_{k}}}$

    ${w_{k+Q} := w_{k+Q} - \eta_{2}\frac{\partial Loss_{2}(w_{k+Q}, B, P_{1})}{\partial w_{k+Q}}}$
\end{center}

where ${P_{1}}$ and ${P_{2}}$ represent the softmax predictions of the weights ${w_{k}}$ and ${w_{k+Q}}$ respectively on the mini-batch data ${B}$. The loss functions ${Loss_{1}}$ and ${Loss_{2}}$ are described below:
\begin{center}
    ${Loss_{1}(w_{k}, B, P_{2}) = L_{c}(P_{1}, y) + D_{KL}(P_{2} || P_{1})}$

    ${Loss_{2}(w_{k+Q}, B, P_{1}) = L_{c}(P_{2}, y) + D_{KL}(P_{1} || P_{2})}$
\end{center}

where ${L_{c}}$ represents the MSE (Mean Squared Error) function, ${y}$ represents the true output of the mini-batch data ${B}$, and ${D_{KL}}$ represents the Kullback-Leibler divergence function. 

The proposed advantages of the \textbf{Def-KT} algorithm are summarised below
\begin{itemize}
    \item Prevents homogenization of different models.
    \item Enhances the generalization ability by synthesizing the knowledge from models with distinct expertise.
    \item Ability to learn from unseen data samples indirectly.
    \item Overcoming catastrophic forgetting.
\end{itemize}

The \textbf{Def-KT} algorithm was evaluated against two other baseline algorithms, \textbf{FullAvg} and \textbf{Combo}. The details of the FullAvg and Combo algorithm are summarised as follows. In both the algorithms, ${Q}$ clients are selected for training out of a total of ${K}$ clients with ${(2Q << K)}$. The ${Q}$ clients train locally on their datasets. In the \textbf{FullAvg} algorithm, the ${Q}$ clients share the updated weights with the next set of ${Q}$ clients, wherein the sets do not overlap. The new set of ${Q}$ clients update their weights to be the average of the shared weights and their existing set of weights. In the \textbf{Combo} algorithm the ${Qth}$ client partitions its weights into two segments and transmits the second half to the ${(Q+K)th}$ client. The ${(Q+K)th}$ client also partitions its current weight into two segments and transmits the first half to the ${Qth}$ client. The ${Qth}$ and ${(Q+K)th}$ clients update their respective weights to be the average of their current weights and the shared weights.

To evaluate the performance of the \textbf{Def-KT} algorithm, a series of experiments were conducted. The details of the experiments are summarised as follows. Four datasets were used for evaluation purposes, namely MNIST, Fashion-MNIST, CIFAR-10, and CIFAR-100. Two types of neural networks namely, MLP and CNN are used depending on the type of dataset used for evaluation. Non-IID data distribution is achieved by distributing only ${\epsilon}$ classes to the respective clients. The smaller the value of ${\epsilon}$, the more heterogeneous the data distribution. For each client, ${0.8}$ fraction of the data is used as the private training data, and ${0.2}$ fraction is used as the local validation set. There is a test dataset that is used to evaluate the global classification accuracy. The models for different clients are initialized with the same model architecture and model parameters. The participation rate is fixed at ${0.2}$ fraction of clients.

The significant findings from the evaluation are summarised below
\begin{itemize}
    \item \textbf{Def-KT} attains a higher global classification accuracy after a fixed number of rounds in most cases when compared with the baseline methods, especially under non-IID data settings.
    \item Global classification accuracy of the baseline methods oscillates more severely than the \textbf{Def-KT} algorithm, especially under non-IID data settings.
    \item \textbf{Def-KT} performs better than the baseline methods irrespective of the number of clients involved in training.
\end{itemize}

\section{Discussion and Conclusions}
In this report, we looked at the challenges that data heterogeneity brings in the paradigm of Federated learning. We also explored a few algorithms that can be used to achieve good accuracy with Federated learning when using heterogeneous and non-IID data. 

In \cite{tao_2020_ensemble}, the author explored the use of knowledge distillation to maintain a global model based on the ensemble of all the participating client models. This ensures the divergence between the output of the global model and the participating client models is minimized. 

In \cite{a}, the author modified the \textbf{FedAvg} algorithm by including weighting parameters based on the number of labels or classes present in the data that a given participating client was trained on, alongside the weighting based on the number of data points that a given participating client was trained on.

In \cite{MKT}, the author explored the use of mutual knowledge transfer in a decentralized federated learning setup to train models on the participating clients simultaneously with an interdependent loss function.

The field of Federated learning is evolving with significant work being done in various aspects ranging from training-related challenges to security-related challenges. Some of the active areas of research in federated learning associated with challenges brought about by data heterogeneity are mentioned below:

\textbf{Heterogeneous and Non-IID Data Handling:}
Handling heterogeneous and non-IID (non-identically distributed) data in federated learning is a significant challenge, as it involves addressing the diversity of data across different clients, which may have distinct data distributions, feature spaces, and task objectives. Some key aspects of heterogeneous and non-IID data handling in the context of federated learning include \textit{model aggregation techniques}, \textit{data transformation and normalization}, \textit{task formulation and decomposition}, \textit{adaptive learning rates}, and \textit{customized model architecture}.

\textbf{Adaptive and Dynamic methods:} 
Adaptive and dynamic methods in federated learning refer to approaches that enable the model and algorithms to adapt to changing conditions and characteristics of the participating clients over time. This adaptability is crucial in scenarios where the distribution of data across clients is not static, and there may be shifts, variations, or other dynamic aspects in the data landscape. Some key aspects of adaptive and dynamic methods in federated learning include \textit{Adaptive learning rates}, \textit{Dynamics model update strategies}, \textit{Transfer learning and continual learning}, \textit{Robustness to client dropouts and joining} and \textit{Decentralized and edge computing}.

\textbf{Sparse and Imbalanced Data:}
Sparse and imbalanced data scenarios in federated learning refer to situations where certain clients have limited data, and the distribution of data across clients is uneven. Dealing with sparse and imbalanced data is essential for achieving fair and effective model performance across all participating clients.  Some key considerations for handling sparse and imbalanced data in federated learning include \textit{sparse data}, \textit{imbalanced data}, \textit{client weighting}, and \textit{transfer learning for sparse clients}.

\textbf{Model Personalization:} 
Model personalization in the context of federated learning refers to the ability to customize or tailor the global model to individual clients' specific characteristics, preferences, or requirements. The goal is to allow each client to benefit from the global knowledge shared by the federation while accommodating local variations in data distributions or user preferences. Some key aspects of model personalization in federated learning include \textit{personalization through hyperparameters}, \textit{dynamic feature importance}, and \textit{personalized federated averaging}.

\textbf{Transfer Learning in federated settings:}
Transfer learning in federated settings involves leveraging knowledge gained from one client or domain to improve the learning or performance of another client or domain within the federated learning framework. The key idea is to transfer learned representations, features, or knowledge from a source task or client to enhance the learning process on a target task or client. Some key aspects of transfer Learning in federated learning include \textit{feature extraction and fine-tuning}, \textit{multi-source transfer learning}, \textit{domain adaptation}, and \textit{cross-client knowledge transfer}.

\textbf{Fairness and Bias Mitigation:}
Fairness and bias mitigation in federated learning are crucial considerations to ensure that machine learning models are fair, and unbiased, and do not discriminate against certain groups or individuals. Addressing fairness concerns is especially important when dealing with data from diverse sources, as biases in training data can lead to biased models. Some key aspects of fairness and bias mitigation in the context of federated learning include \textit{fair representation}, \textit{fair aggregation}, \textit{bias detection and measurement}, and \textit{demographic parity and equalized odds}.

\textbf{Incorporating Domain Knowledge:}
Incorporating domain knowledge in federated learning refers to the integration of expert knowledge or prior information about the problem domain into the learning process. This knowledge can help improve model performance, generalization, and interpretability. Some key aspects of incorporating domain knowledge in the context of federated learning include \textit{feature engineering}, \textit{task-specific customization}, \textit{constraint incorporation}, and \textit{dynamic learning rate adjustment}.

\nocite{*}
\printbibliography
\end{document}